\documentclass{article} 
\usepackage{iclr2025_conference,times}

\usepackage{amsmath,amsfonts,bm}

\def\eqref#1{equation~\ref{#1}}

\def\1{\bm{1}}

\DeclareMathAlphabet{\mathsfit}{\encodingdefault}{\sfdefault}{m}{sl}
\SetMathAlphabet{\mathsfit}{bold}{\encodingdefault}{\sfdefault}{bx}{n}

\usepackage{xparse}
\NewDocumentCommand{\bang}{ mO{} }{\textcolor{blue}{\textsuperscript{\textit{bang}}\textsf{\textbf{\small[#1]}}}}

\usepackage{url}

\usepackage{tabularray}
\usepackage{color}
\usepackage{booktabs}
\usepackage{siunitx}
\usepackage{graphicx}
\usepackage{array}

\usepackage{listings}
\usepackage{xcolor}
\usepackage{algorithm}
\usepackage{algorithmic}
\usepackage{wrapfig}
\usepackage{tabularx}
\usepackage{subcaption}
\usepackage{pifont}
\newcommand{\cmark}{\ding{51}}%
\newcommand{\xmark}{\ding{55}}%
\usepackage{caption}
\usepackage[hyperfootnotes=false]{hyperref}
\lstdefinestyle{pythonstyle}{
    language=Python,
    backgroundcolor=\color{white},
    basicstyle=\ttfamily\scriptsize,
    keywordstyle=\color{blue},
    commentstyle=\color{gray},
    stringstyle=\color{red},
    numbers=left,
    numberstyle=\tiny,
    stepnumber=1,
    numbersep=5pt,
    frame=single,
    rulecolor=\color{black},
    tabsize=4,
    breaklines=true,
}

\lstdefinestyle{txtfile}{
    basicstyle=\scriptsize\ttfamily,
    backgroundcolor=\color{white},
    frame=single,
    rulecolor=\color{black!30},
    stringstyle=\color{red},
    numbers=left,
    numberstyle=\tiny\color{gray!50},
    numbersep=5pt,
    breaklines=true,
    breakatwhitespace=true,
    showstringspaces=false,
    tabsize=4,
    captionpos=b
}

\title{\methodname{}: Tree-Search Enhanced LLM Agents for Automated Machine Learning}

\author{
    \textbf{Yizhou Chi}$^{1,2} \thanks{These authors contributed equally to this work.}$,
    \textbf{Yizhang Lin}$^1 \footnotemark[1]$,
    \textbf{Sirui Hong}$^1$,
    \textbf{Duyi Pan}$^3$,
    \textbf{Yaying Fei},
    \textbf{Guanghao Mei}$^4$, \\
    \textbf{Bangbang Liu}$^1$,
    \textbf{Tianqi Pang}$^5$,
    \textbf{Jacky Kwok}$^6$,
    \textbf{Ceyao Zhang}$^7$,
    \textbf{Bang Liu}$^8 \thanks{Bang Liu (E-mail: bang.liu@umontreal.ca) and Chenglin Wu (E-mail: alexanderwu@deepwisdom.ai) are the corresponding authors.}$,
    \textbf{Chenglin Wu}$^1 \footnotemark[2]$
    \vspace{1.5em} 
    \\
    $^1$DeepWisdom, 
    $^2$University of California, Berkeley, \\ 
    $^3$The Hong Kong University of Science and Technology (Guangzhou), \\
    $^{4}$University of California, San Diego, 
    $^5$South China Normal University, \\
    $^6$Stanford University, 
    $^{7}$The Chinese University of Hong Kong, Shenzhen, \\
    $^{8}$Université de Montréal \& Mila
}

\newcommand{\methodname}{\textsc{SELA}}

\iclrfinalcopy 
\begin{document}

\maketitle


\begin{abstract}

Automated Machine Learning (AutoML) approaches encompass traditional methods that optimize fixed pipelines for model selection and ensembling, as well as newer LLM-based frameworks that autonomously build pipelines. While LLM-based agents have shown promise in automating machine learning tasks, they often generate low-diversity and suboptimal code, even after multiple iterations. To overcome these limitations, we introduce Tree-\textbf{S}earch \textbf{E}nhanced \textbf{L}LM \textbf{A}gents (\textbf{\methodname{}}), an innovative agent-based system that leverages Monte Carlo Tree Search (MCTS) to optimize the AutoML process. By representing pipeline configurations as trees, our framework enables agents to conduct experiments intelligently and iteratively refine their strategies, facilitating a more effective exploration of the machine learning solution space.
This novel approach allows \methodname{} to discover optimal pathways based on experimental feedback, improving the overall quality of the solutions. In an extensive evaluation across 20 machine learning datasets, we compare the performance of traditional and agent-based AutoML methods, demonstrating that \methodname{} achieves a win rate of 65\% to 80\% against each baseline across all datasets. These results underscore the significant potential of agent-based strategies in AutoML, offering a fresh perspective on tackling complex machine learning challenges\footnote{The code is available at \url{https://github.com/geekan/MetaGPT}}.

\end{abstract}

\section{Introduction}

Automated Machine Learning (AutoML) is a rapidly evolving field that seeks to automate the process of designing reliable machine learning solutions with minimal human intervention. Traditional AutoML frameworks, such as Auto-WEKA \citep{thornton2013auto}, Auto-Sklearn \citep{feurer-neurips15a, feurer-arxiv20a}, AutoGluon \citep{tang2024autogluon}, and H2O AutoML \citep{H2OAutoML20}, rely on predefined search spaces and routines. These frameworks primarily focus on optimizing hyperparameters and model ensembling to find the best model configuration. However, this fixed and static approach often lacks the adaptability needed to handle diverse and dynamic data scenarios, resulting in suboptimal performance in more complex settings. Additionally, the traditional focus on model training leaves other crucial stages of the machine learning pipeline, such as data preprocessing and feature engineering, underexplored, thereby limiting the overall effectiveness of these systems.

Recently, large language model (LLM)-based agents have emerged as promising tools for automating machine learning tasks by leveraging natural language processing capabilities to generate code. These systems typically begin with a natural language prompt describing the dataset and the problem, after which an LLM generates an end-to-end solution. Early efforts, such as \cite{zhang2024mlcopilotunleashingpowerlarge}, experimented with prompting LLMs to generate machine learning solutions, while \cite{hong2024datainterpreterllmagent} introduced agents equipped with Hierarchical Graph Modeling and Programmable Node Generation to address complex and dynamic workflows. Despite these advances, LLM-based solutions often fall short in generating diverse and highly optimized workflows, as their search process remains limited to a single pass or trial. Without iterative refinement or the ability to explore alternative strategies, these solutions frequently converge on suboptimal results, even when multiple attempts are allowed.

A critical shortcoming of both traditional AutoML and LLM-based frameworks lies in their inability to mimic the nuanced problem-solving approach of human experts. When approaching a machine learning task, an expert does not simply execute a fixed pipeline or rely on a single attempt. Instead, they explore various potential configurations, systematically conduct experiments, analyze results, and iteratively refine their understanding of each component’s effectiveness. This iterative, feedback-driven process allows experts to explore diverse solutions and improve them incrementally until they arrive at the optimal configuration.

\begin{figure}[!thb]
    \centering
    \includegraphics[width=0.98\linewidth]{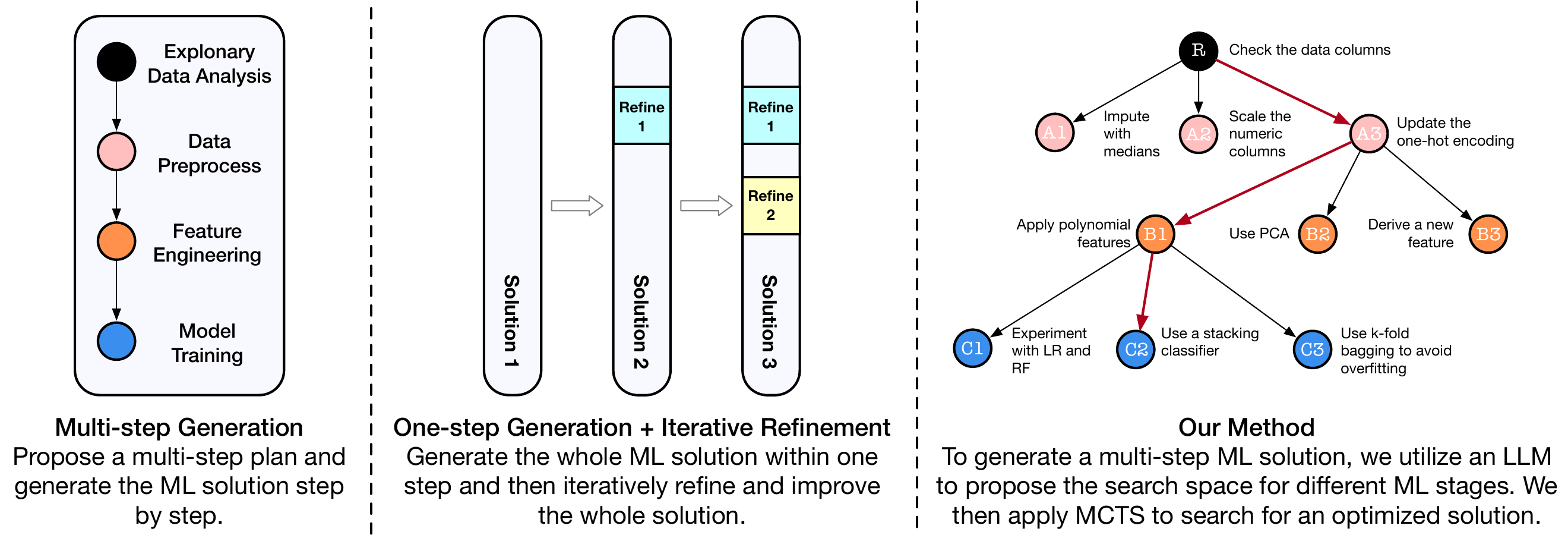}
    \caption{\methodname{}'s abstraction compared to other agent-based AutoML frameworks. There are two main types of agent-based approaches to AutoML problems. The first approach \citep{hong2024datainterpreterllmagent} divides a machine learning task into multiple stages, proposing a plan for each stage, and generating and executing code step by step according to the plan, with no refinement after the solution is completed. The second \citep{Schmidt_Wu_Jiang} generates the entire solution in one step and iteratively refines it as a whole. \methodname{} integrates both approaches, enabling stage-wise planning while iteratively exploring better solutions at each stage level.}
    \label{fig:main-compare}
\end{figure}

Inspired by this human-centered approach, we propose Tree-\textbf{S}earch \textbf{E}nhanced \textbf{L}LM \textbf{A}gents (\textbf{\methodname{}}) for automated machine learning, a novel framework that integrates the strengths of LLM agents with a structured search and refinement process modeled on how experts solve machine learning problems. As illustrated in Figure~\ref{fig:main-compare}, our framework combines the benefits of stage-wise planning, where each stage (e.g., Exploratory Data Analysis, Data Preprocessing, Feature Engineering, and Model Training) is handled sequentially, with an iterative refinement mechanism. 

In \methodname{}, the search space of a machine learning problem is proposed and conceptualized as a tree, where each branch represents a potential solution path. To navigate this search space, we employ Monte Carlo Tree Search (MCTS) \citep{MCTS} as the core decision-making engine, leveraging its ability to balance exploration (testing new strategies) and exploitation (improving known good strategies). MCTS allows the agent to efficiently explore large decision spaces, collect and process experimental results, and intelligently select the next promising configuration to test on. By iterating through this cycle of experimentation and refinement, \methodname{} incrementally improves its solutions, much like an expert who tests and improves its strategy based on continuous feedback.

We rigorously evaluated \methodname{} using 20 diverse datasets from the AutoML Benchmark \citep{JMLR:v25:22-0493}, comparing its performance against both traditional AutoML systems and agent-based AutoML approaches. The results demonstrate that \methodname{} consistently delivers superior performance across a wide range of machine learning tasks, validating its effectiveness and adaptability.

To summarize, our research makes the following contributions: 
\begin{enumerate}

\item We introduce a feedback-driven approach for LLM agents to iteratively explore machine learning configurations, optimizing solutions over multiple experimental rounds.


\item Using Monte Carlo Tree Search, our system navigates a tree-structured search space, adaptively identifying high-performance pipelines through feedback.


\item We compare agent-based and traditional AutoML, highlighting agentic methods’ flexibility and potential for enhanced performance in machine learning.

\end{enumerate}

\begin{table}[H]
    \centering
    \resizebox{0.99\textwidth}{!}{
    \Large 
    \begin{tabular}{lccccc}
        \toprule
        & \shortstack{\textbf{Dynamic}\\\textbf{Pipeline}} & \shortstack{\textbf{Feature}\\\textbf{Engineering}} & \shortstack{\textbf{Model}\\\textbf{Training}} & \shortstack{\textbf{Model}\\\textbf{Improvement}} & \shortstack{\textbf{Pipeline}\\\textbf{Optimization}}\\
        \midrule
        AutoGluon \footnotesize \citep{erickson2020autogluontabularrobustaccurateautoml}  & \xmark  & \xmark & Fixed models & Multi-layer stacking + bagging  & \xmark  \\
        AutoSklearn \footnotesize \citep{feurer-arxiv20a} & \xmark  & \xmark & Fixed models & Bayes Opt. + meta-learning + ensemble & \xmark \\
        Data Interpreter \footnotesize \citep{hong2024datainterpreterllmagent} & \cmark & Instinctive & Instinctive   & Instinctive & \xmark \\
        AIDE \footnotesize \citep{Schmidt_Wu_Jiang} & \cmark & Instinctive &  Dynamic \& diverse & Dynamic \& diverse & One-step refinement + LLM \\
        \methodname{} \footnotesize (Ours) & \cmark  & Dynamic \& diverse &  Dynamic \& diverse & Dynamic \& diverse & Stepwise MCTS + LLM\\
        \bottomrule
    \end{tabular}
    }
    \caption{\normalsize Comparison of key capabilities across various AutoML methods. \textit{Dynamic} indicates the system's ability to adjust workflows based on intermediate outcomes, allowing it to adapt as new information emerges. \textit{Diverse} refers to employing multiple strategies or methods across tasks, which helps capture varied modeling needs. \textit{Instinctive} means that the system directly relies on the decisions generated by an LLM and heavily depends on the model's inclination.}
    \label{table:baseline-comparison}
\end{table}

\section{Related Works}

\noindent\textbf{Tree Search and Its Integration with LLMs}
Tree search algorithms have significantly advanced problem-solving in artificial intelligence, with Monte Carlo Tree Search (MCTS) emerging as a leading technique. These algorithms have been successfully applied across various domains, including robotics~\citep{wu2015, clary2018, graeme2019}, chemistry~\citep{segler2018}, and gaming~\citep{Silver2016MasteringTG, Silver2017MasteringTG}, where MCTS is used to navigate vast solution spaces and solve complex problems. More recently, research has focused on integrating tree search with Large Language Models (LLMs) to enhance reasoning and decision-making. Studies such as \citet{krishnamurthy2024largelanguagemodelsexplore} and \citet{dwaracherla2024efficientexplorationllms} explored LLMs’ capacities for efficient exploration, while \citet{tang2024coderepairllmsgives} and \citet{hui2024rotenhancinglargelanguage} developed strategies for exploiting previously learned knowledge. \citet{zhou2024languageagenttreesearch} and \citet{chi2024thoughtsculptreasoningintermediaterevision} applied MCTS for planning with external or self-evaluated feedback, while \citet{feng2023alphazero, wang2024litesearchefficacioustreesearch} adapted AlphaZero-style tree search to LLM-based tasks. These advancements underscore the potential of combining tree search methods with LLMs, balancing exploration of new solutions with exploitation of prior knowledge to enhance decision-making.


\noindent\textbf{Advances and Limitations in AutoML Systems}
Automated Machine Learning (AutoML) frameworks were introduced to reduce the need for expert knowledge in designing machine learning pipelines. Early AutoML efforts, such as \citep{thornton2013auto, olson2016tpot, jin2019auto, feurer-arxiv20a, erickson2020autogluontabularrobustaccurateautoml, H2OAutoML20, wang2021flaml}, focused primarily on automating key pipeline components like hyperparameter optimization, model selection, stacking, and ensembling. These frameworks achieved notable progress by integrating meta-learning and hyperparameter search strategies to automatically select and tune machine learning models. Furthermore, extensions into multi-modal data settings~\citep{tang2024autogluon, jin2023autokeras} have broadened AutoML’s applicability.

Recently, there has been growing interest in leveraging LLMs within AutoML systems to enhance pipeline flexibility. Studies such as \cite{hollmann2024large, li2024exploringlargelanguagemodels} applied LLMs to automate feature engineering, while \citet{liu2024large} introduced LLMs for hyperparameter tuning. In addition, \citet{luo2024autom3l} proposed embedding LLMs at each stage of the machine learning workflow. Despite these advancements, traditional AutoML systems remain constrained by rigid pipelines and limited flexibility to adapt to unique datasets or specific task requirements.

\noindent\textbf{LLM Agents for Dynamic Machine Learning Pipelines}
In contrast to static pipelines, LLM-based agents offer a more dynamic solution for addressing complex machine learning challenges. \citet{ hong2024datainterpreterllmagent, hong2024metagpt} introduced an LLM agent with hierarchical graph modeling and programmable node generation, enabling the creation of sophisticated, adaptable pipelines for diverse data scenarios. Similarly, \citet{zhang2024mlcopilotunleashingpowerlarge} demonstrated that LLMs could effectively interpret structured inputs and apply past experiences to solve new machine learning tasks. \citet{guo2024dsagentautomateddatascience} expanded on this by introducing a data science agent that leverages case-based reasoning; however, it faces challenges when generating solutions from scratch due to its reliance on existing codebases. \citet{Schmidt_Wu_Jiang} proposed an iterative approach, where the entire pipeline is generated in one step and refined iteratively through incremental modifications.

Building on these efforts, \methodname{} introduces an agent that integrates the strengths of both approaches—stage-wise planning and iterative refinement—allowing it to autonomously explore and generate machine learning solutions from the ground up. This approach offers greater flexibility and control during the search process, enabling the generation of optimized solutions at each stage. Table \ref{table:baseline-comparison} highlights the functionalities provided by different AutoML systems.


\section{Method}

\begin{figure}[!t]
    \centering
    \includegraphics[width=0.98\linewidth]{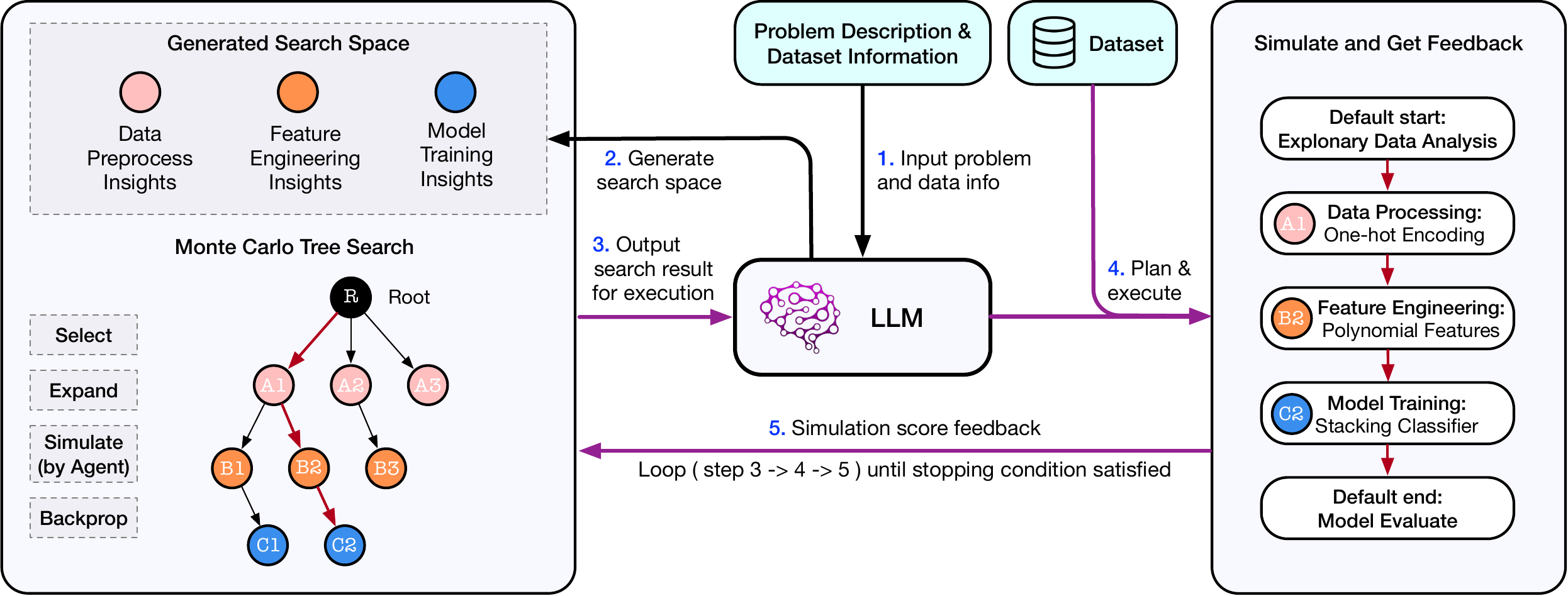}
    \caption{\methodname{}’s pipeline operates as follows: The system begins by inputting the problem description and dataset information into the LLM, which generates a search space of potential solutions, encompassing data preprocessing, feature engineering, and model training. The search module, powered by Monte Carlo Tree Search (MCTS), explores this space by selecting, expanding, and simulating potential configurations. The LLM agent then simulates the selected configuration by planning, coding, and executing the experiment. Feedback from the simulation is fed back into the search module, where it is used in the backpropagation step to refine future searches. This iterative process continues until a predefined stopping criterion is met, resulting in an optimized experimental pipeline.}
    \label{fig:pipline}
\end{figure}

As illustrated in Figure~\ref{fig:pipline}, \methodname{} consists of three key components: an LLM-based insight proposer, a search module using MCTS, and an LLM agent as the experiment executor. First, the LLM generates insights from the problem description and dataset, defining a search space. The search module then organizes this space into a tree structure and uses MCTS to explore promising paths. During each cycle, the selected path is passed to the LLM agent, which translates the configuration into an executable pipeline. The agent plans, codes, and executes the experiment, feeding the results back to refine future searches. This iterative process continues until the termination criterion is met. The following sections provide a detailed explanation of each component.

\subsection{Insight Proposal and Search Space Creation}
To enable \methodname{} to explore a wide range of machine learning strategies, we introduce an insight proposer that generates diverse methods tailored to different stages of the machine learning workflow. Each proposed insight suggests either a single technique or a combination of methods aimed at enhancing performance. For instance, a feature engineering insight might recommend creating interaction features from existing variables, while a model training insight could propose a specific algorithm or suggest running a grid search to improve accuracy.

The insight proposer takes as input the problem description $p$ and dataset information $d$, such as metadata and sample records, and generates $m$ insights $\lambda$ for each stage of the machine learning process using a large language model $M$. These insights are stored in an insight pool, forming a search space $\Lambda$ for \methodname{} to explore. We decompose the machine learning process into five stages: Exploratory Data Analysis ($\tau_1$), Data Preprocessing ($\tau_2$), Feature Engineering ($\tau_3$), Model Training ($\tau_4$), and Model Evaluation ($\tau_5$). For simplicity, we denote the entire set of stages as $T$ and refer to any specific stage as $\tau$.
\begin{align}
    \text{InsightProposer}(p, d, M) \rightarrow \Lambda := \{\lambda_i^\tau \mid \tau \in T, i = 1, \dots, m \}
\end{align}

\subsection{Pipeline Execution and Code Generation}
We employ an LLM agent, referred to as the experiment executor $E$, to conduct each trial by building practical experimental pipelines from natural language requirements. The agent takes two main steps in this process. First, given an experiment configuration $c$, which is a set of insights provided by the search module (introduced in Section \ref{section:tree-search}), the experiment executor translates these insights into a detailed plan. This plan consists of a sequence of task instructions $I^{\tau \in T}$ corresponding to each stage of the machine learning process. This step is referred to as $E_{\text{plan}}$.

Next, following the plan, the agent writes and executes code $\sigma^\tau$ for each task $\tau$ based on the respective instruction $I^\tau$, producing the code $\sigma^{\tau \in T}$ for the full pipeline, along with the final execution score $s$. The complete set of code outputs $\sigma^{\tau \in T}$ is concatenated into a full solution $\sigma_{sol}$ to address the problem. This phase is referred to as $E_{\text{code \& execute}}$.
\begin{align}
    E_{\text{plan}}(p, d, c, M) & \rightarrow I^{\tau \in T} \\
    E_{\text{code \& execute}}(I^{\tau \in T}, D, M) & \rightarrow (\sigma^{\tau \in T}, s)
\end{align}

\subsection{Tree Search in Machine Learning Experiments}
\label{section:node-exploration}

In order to systematically explore the different configurations in machine learning experiments, we model the search space as a hierarchical tree. This structure allows us to apply tree search algorithms, where each path through the tree represents a different experiment configuration. Algorithm \ref{code} also provides an overview of this searching process.

\subsubsection{Experiment Node}
\label{section:node}

To facilitate the exploration of various strategies, we model the proposed search space as a hierarchical tree that is well-suited for applying search algorithms. Each node in the tree, denoted as $x$, represents one insight $\lambda$ in the search space $\Lambda$ and contains the following attributes:

\begin{itemize}
    \item \textbf{Insight} $\lambda(x)$: Represents the specific insight $\lambda_i^\tau \in \Lambda$ associated with this node, where $\tau$ denotes the stage of the machine learning pipeline.
    \item \textbf{Depth} $\delta(x)$: Indicates the stage of the machine learning process the node corresponds to (e.g., depth 1 might represent data preprocessing, depth 2 for feature engineering, and depth 3 for model training).
    \item \textbf{Value} $v(x)$: The cumulative score from simulations for this node and all its descendants.
    \item \textbf{Number of Visits} $n_{\text{visits}}(x)$: The total number of simulations conducted for this node and its descendants.
    \item \textbf{Simulation Score} $s(x)$: The score for simulating this node.
    \item \textbf{Solution Code} $\sigma_{\text{sol}}(x)$ The final code produced after the node simulation.
    \item \textbf{Stage Code} $\sigma_{\text{stage}}(x)$: The code generated up to the node's current stage, a part of the solution code
\end{itemize}

By modeling the search space as a tree, each path from the root to a node $x$ represents an experiment configuration $c(x) = \{\lambda(x_1), \lambda(x_2), \dots, \lambda(x)\} \subset \Lambda$, where $x_1, x_2, \dots, x$ are nodes along the path. The task of finding the optimal solution can therefore be viewed as a path search within the tree, where each path corresponds to a potential configuration of the experiment.
\subsubsection{Tree Search for ML Experiments}
\label{section:tree-search}

We apply Monte Carlo Tree Search (MCTS) to systematically explore and identify optimal machine learning solutions within our framework. MCTS allows us to efficiently navigate the search space across multiple stages of the machine learning pipeline—from data preprocessing to model selection—by balancing exploration and exploitation.

\begin{algorithm}[H]
    \caption{\methodname{} using MCTS}
    \begin{algorithmic}[1]
        \REQUIRE Problem description $p$, data information $d$, data $D$, LLM $M$, rollout number $k$.
        
        \STATE $\Lambda \leftarrow \text{InsightProposer}(p, d, M) $
        \STATE Initialize Tree using $\Lambda$
        \FOR{$i$ = 1 \textbf{to} $k$}
            \STATE node $x \leftarrow$ select(Tree)
            \STATE $X_{\text{child}}\leftarrow$ expand(Tree, $x$)
            \STATE Randomly sample a node $x_{\text{sample}}$ from $X_{\text{child}}$
            \STATE Retreive experiment configuration $c(x_\text{sample})$
            \STATE $\sigma_{sol}, s \leftarrow \text{simulate}(c(x_\text{sample}), p, d, D, M)$
            \STATE attach the simulation result $\sigma_{sol}, s$ to $x_\text{sample}$ for final solution selection
            \STATE Backpropagate(Tree, $s$)
        \ENDFOR
        \STATE $x_{\text{dev best}} \leftarrow \underset{x \in \text{Tree}}{\text{argmax}}(s(x))$

        \ENSURE $\sigma_{sol}(x_{\text{dev best}})$
    \end{algorithmic}
    \label{code}
\end{algorithm}

\begin{algorithm}[H]
    \caption{Simulate}
    \begin{algorithmic}[1]
        \REQUIRE Experiment configuration $c$, problem description $p$, data information $d$, data $D$, LLM $M$.
        \STATE Draft plans $I^{\tau \in T} \leftarrow E_{\text{plan}}(p, d, c, M)$
        \STATE Code and execute sequentially $\sigma^{\tau \in T}, s \leftarrow E_{\text{code \& execute}}(I^{\tau \in T}, D, M)$
        \STATE $\sigma_{sol} \leftarrow \text{concatenate}(\sigma^{\tau \in T})$

        \ENSURE $\sigma_{sol}, s$
    \end{algorithmic}
    \label{simulate-code}
\end{algorithm}

The search process involves performing multiple rollouts, which include the steps of selection, expansion, simulation, and backpropagation. We conduct $k$ rollouts to explore various paths, aiming to identify the best solution.

\noindent\textbf{Selection}
At each iteration, we use a modified version of the UCT (Upper Confidence Bound for Trees) algorithm \citep{10.1007/11871842_29}, referred to as UCT-DP (depth-preferred), to select a node from the search tree. Unlike traditional MCTS, where simulations are often performed quickly due to a fixed action space and negligible action time, the context of machine learning tasks presents a different challenge. Processes such as model training introduce significant computational time, making efficient node exploration crucial. Since model selection can heavily influence the overall machine learning performance, we prioritize exploring nodes at greater depths early on.

This modification reduces the need to explore every unvisited node, allowing deeper nodes to be reached in fewer iterations—making the approach better suited for large-scale machine learning experiments. The modified selection algorithm is expressed as:

\begin{align}
\label{eq:sel_alg}
\text{UCT-DP}(x) = \frac{v(x)}{n(x)} + \alpha_{\text{explore}} \sqrt{\frac{\ln n_{\text{visits}}(x_{\text{parent}})}{n(x)}}
\end{align}

\begin{align}
    n(x) = \begin{cases}
        \alpha_{\text{unvisted}} & \text{if } n_{\text{visits}}(x) = 0 \\
        n_{\text{visits}}(x) & \text{otherwise.}
    \end{cases}
\end{align}

Here, $\alpha_{\text{unvisted}}$ is a constant between 0 and 1 controlling the selection preference for unvisited nodes, balancing between full exploration and computational efficiency. This adjustment allows us to focus more on deeper parts of the tree that are likely to yield better solutions. 

\noindent\textbf{Expansion}
During the expansion phase, a set of child nodes $X_{\text{child}}$ are instantiated from the selected node $x$ for potential simulation. Note that a child node $x_{\text{child}}$ from the node $x$ at depth $\delta$ inherits the attributes of $x$ and possesses $\lambda(x_\text{child}) \rightarrow \lambda^{\tau_{\delta+1}}$, an insight of stage $\tau_{\delta+1}$ from the search space.





\noindent\textbf{Simulation}
Once expanded, a node $x_{\text{sample}}$ is randomly sampled from $X_{\text{child}}$ for simulation. The path from root to the sampled node forms a set of insights $c(x_\text{sample}) = \{\lambda(x_1),\lambda(x_2),...,\lambda(x_\text{sample})\} \subset \Lambda$, representing the experiment configuration to be simulated, where $x_1,x_2,..,x_\text{sample}$ are the nodes along the path. The configuration $c(x_\text{sample})$ is then fed to the experimenter $E$ for execution following $E_\text{plan}$ and $E_\text{code \& execute}$, which produces a simulation score $s$, as illustrated in Section \ref{section:node}. The score serves as the feedback for back propagation. Algorithm \ref{simulate-code} outlines the simulation process.



\noindent\textbf{Backpropagation}
After the simulation concludes, the performance score (e.g., based on the development set) is retrieved and backpropagated through the tree. The score is propagated from the simulated node up to the root, updating each parent node's value and visit count. This allows nodes representing more promising solutions to be prioritized in future rollouts. In addition, the solution code is also backpropagated up to the tree, and it can be processed and saved as stage code depending on the parent node during the update.

Backpropagation ensures that the algorithm learns which paths yield better results, guiding the search toward higher-performing nodes as more rollouts are conducted.

\subsubsection{Experiment State Saving and Loading}
To boost experimentation efficiency and reduce token usage, \methodname{} implements fine-grained code reuse by caching code at the stage level for each attempted configuration $c$. This allows the framework to reuse as much saved code as possible when a new configuration $c_\text{new}$ shares components with existing ones. Additionally, this technique addresses the challenge of LLM non-determinism, where identical instructions can produce different code, increasing variance in final performance. Specifically, whenever a node is chosen for execution, the experimenter loads and reruns the saved stage code, if available, ensuring consistency before progressing to the next stage. This approach effectively conserves resources while maintaining robust performance across stages. In Appendix \ref{section:cost}, we examine the cost efficiency of this state-saving and loading mechanism.
\section{Experiments}

\subsection{Experimental Setup}

\paragraph{Datasets}
We evaluate \methodname{} alongside several baselines on 20 datasets, which include 13 classification tasks and 7 regression tasks from the AutoML Benchmark (AMLB) \citep{JMLR:v25:22-0493} and Kaggle Competitions.


Table \ref{table:datasets} provides detailed information on the datasets used. All datasets are split into training, validation, and test sets with a 6:2:2 ratio. Each framework utilizes the training and validation sets to train models and makes predictions on the test set labels.



\paragraph{Evaluation Metrics}
For the AMLB datasets, we use the default target column provided by OpenML. For Kaggle competition datasets, we rely on the target column specified in the competition description. Performance is measured using root mean squared error (RMSE) for regression tasks, F1 score for binary classification, and F1-weighted score for multi-class classification. To ensure comparability across datasets with varying metrics, we introduce a Normalized Score (NS), which maps RMSE into the range from 0 to 1.
\begin{align} 
    \text{NS}(s_{\text{raw}}) = \begin{cases} 
        \frac{1}{1 + \log{(1 + s_{\text{raw}})}} & \text{if the metric is RMSE.} \\
        s_{\text{raw}} & \text{otherwise.}
    \end{cases} 
\end{align}

Here, $s_{raw}$ represents the raw score before normalization.
To evaluate \methodname{} against other frameworks, we employ three key metrics: average Normalized Score (NS), average rank, and average best rank. The average rank is calculated by considering all rankings of a method across datasets, while the average best rank focuses on the method's best performance in each dataset.
We also want to quantify how other baselines perform relative to \methodname{}. The ``Rescaled NS" is defined as:
\begin{align}
    \text{Rescaled NS}(f) = \frac{\text{NS}_{f}}{\text{NS}_{\methodname{}}}
\end{align}
where $f$ represents the baseline method being compared to \methodname{}.

\paragraph{Method and Baselines Setup}
We compare \methodname{} with several baseline methods, including Data Interpreter \citep{hong2024datainterpreterllmagent}, AIDE \citep{Schmidt_Wu_Jiang}, AutoGluon \citep{erickson2020autogluontabularrobustaccurateautoml}, and AutoSklearn \citep{feurer-neurips15a, feurer-arxiv20a}.

For our LLM-based approaches (\methodname{}, Data Interpreter, and AIDE), we employ a consistent initial task prompt across all methods. This prompt encompasses the dataset name, target column, and evaluation metric. We choose DeepSeek v2.5 \citep{deepseekv2} as our foundation LLM due to its open-source nature, strong coding capabilities, and cost-effective token usage.
To encourage output diversity, we set the temperature parameter to 0.5 for all LLM-based methods. AIDE conducts 10 iterations per execution, while \methodname{} performs 10 rollouts. 

For \methodname{}, we employ Data Interpreter as the experimenter, leveraging its multi-step generation capability. We configured the hyperparameters of UCT-DP as follows: $\alpha_{\text{unvisited}}$ is set to 0.8 and $\alpha_{\text{explore}}$ is set to 1.4. These settings aim to balance exploration and exploitation in the method's search strategy.

Each method, except for AutoGluon, is run three times for each dataset. AutoGluon, being deterministic, is run only once with its default settings. AutoSklearn is also run with default settings, limited to 600 seconds per task.

\begin{table}[hbt]
\centering
\resizebox{1.0\textwidth}{!}{
    \begin{tabular}{lccccccc}
        \toprule
        \textbf{Method} &  \textbf{Wins} & \textbf{Losses} & \textbf{Top 1}  & \textbf{Avg. NS} \% $\uparrow$ & \textbf{Avg. Best NS} \% $\uparrow$ & \textbf{Avg. Rank} $\downarrow$ &  \textbf{Avg. Best Rank} $\downarrow$ \\
        \midrule
        AutoGluon & 7 & 13  & 4 & 53.2 & 53.2 & \textbf{4.4} & 4.4\\
        AutoSklearn & 5  & 15 & 5 & 46.1 & 47.5  & 7.6 & 6.1\\
        AIDE & 5 & 15 & 2 & 47.1 & 51.8 & 7.8 & 5.3 \\
        Data Interpreter & 4 & 16 & 2 & 47.4 & 50.2 & 8.8 & 6.4\\
        \methodname{} & - & - & \textbf{7} & \textbf{53.3} & \textbf{54.7} & 4.8 & \textbf{2.7}\\
        \bottomrule
    \end{tabular}
    }
    \caption{Results of each AutoML framework on 20 tabular datasets. The ``Wins" column indicates the number of datasets where the method outperforms \methodname{}, while ``Losses" shows the number of datasets where the method underperforms. The ``Top 1" column represents the number of datasets where the method produces the best predictions across methods.}
    \label{table:main}
\end{table}

\subsection{Results}

\begin{figure}[hbt]
    \centering
    \includegraphics[width=0.95\linewidth]{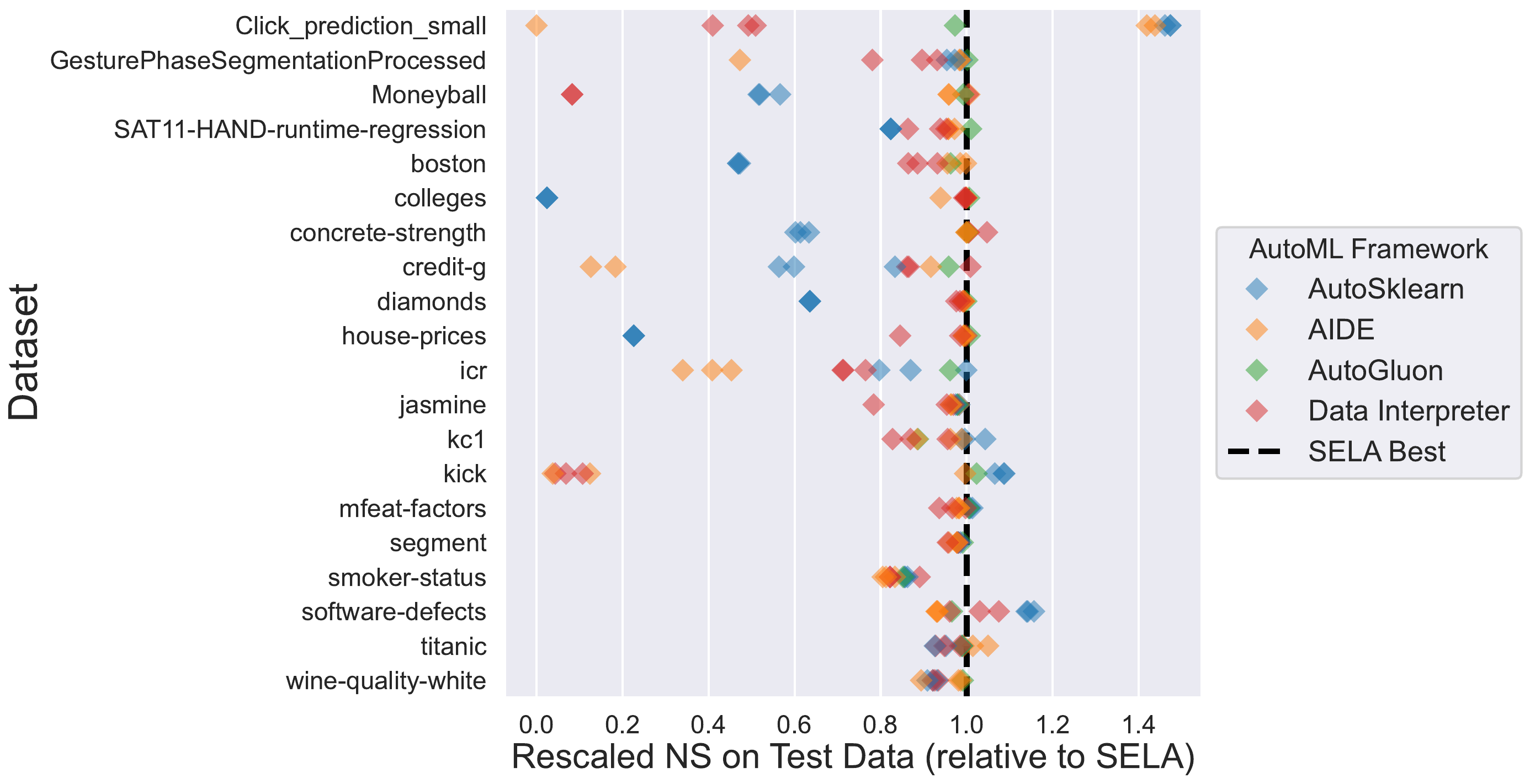}
    \caption{Rescaled NS of AutoML frameworks relative to \methodname{} on tabular datasets. Points to the left of the vertical line indicate poorer predictions compared to \methodname{}. Notably, \methodname{} often occupies a leading position across the datasets.}
    \label{fig:main-scatter}
\end{figure}

As shown in Table \ref{table:main}, \methodname{} achieves the highest average Normalized Score (NS) and average best rank among all frameworks. Notably, \methodname{} excels in producing the highest number of top predictions, as indicated in the ``Top 1" column across all datasets. Furthermore, the ``Losses" column reveals that each competing method falls short against \methodname{}, losing in 65-80\% of the datasets.

Interestingly, AutoGluon exhibits a marginally higher average rank than \methodname{}. This slight discrepancy may be attributed to the inherent randomness in LLMs and model training processes, which can influence the exploration of machine learning solutions. However, \methodname{}'s higher average NS suggests that it performs strongly in the datasets where it excels, while its losses in other datasets are relatively minor. This means that even when \methodname{} produces lower-ranked solutions, the performance gap is small, allowing it to fully compensate in the datasets where it performs well.

The two other agent-based methods exhibit relatively lower performance. The first method, Data Interpreter, struggles to enhance its score with multiple attempts due to its inability to refine its solution after completing a machine learning task. The second method, AIDE, does not have a stage-specific planning module, limiting its capacity to improve results after a series of greedy exploitation, which makes it prone to falling into local optima. These limitations likely account for their weaker performance.

Figure \ref{fig:main-scatter} further corroborates \methodname{}'s effectiveness, revealing that its best solutions frequently occupy leading positions across various datasets. This visual representation exhibits the method's consistent high performance and adaptability across different ML datasets. We also include a detailed results of each method in Appendix \ref{appendix:results}.


\subsection{Ablation Study}
For the rest of the study, we employ a subset of datasets to evaluate \methodname{} under various settings. Our selection process involves choosing the first two datasets alphabetically for each machine learning task. Specifically, we use boston, colleges, credit-g, Click\_prediction\_small, GesturePhaseSegmentationProcessed, and mfeat-factors to conduct the ablation study.

\begin{table}[hbt]
\centering
\begin{tabular}{lccc}
    \toprule
    & \textbf{Data Interpreter} & \textbf{\methodname{} (Random Search)} & \textbf{\methodname{} (MCTS)}  \\
    \midrule
    Avg. NS $\uparrow$ & 56.4    & 58.6       & \textbf{60.9} \\
    Avg. Best NS $\uparrow$ & 59.0     & 61.4        &  \textbf{62.4} \\
    \midrule
    Avg. Rank $\downarrow$        & 6.9     & 4.8     &  \textbf{3.3}    \\
    Avg. Best Rank $\downarrow$     & 4.8      & 2.8     & \textbf{1.5}    \\ 
    \bottomrule
\end{tabular}
\caption{Performance results for each search setting on the chosen datasets. \methodname{} with MCTS consistently surpasses \methodname{} with Random Search.}
\label{table:tree}
\end{table}

\paragraph{Effectiveness of Search}
To evaluate the effectiveness of Monte Carlo Tree Search (MCTS) in improving the solution search process, we conducted an ablation study. In this study, we compared the performance of our method using MCTS against a variant that randomly samples insights from each stage's insight pool.
As shown in Table \ref{table:tree}, the MCTS version achieves a higher average normalized score across datasets and a better overall ranking compared to the random sampling approach. Moreover, even the random sampling variant of our method outperforms Data Interpreter, the base experimenter. This suggests the presence of an appropriate search space and an experiment agenda is vital for improving a machine learning agent. Our insight proposer generates relevant and useful insights, facilitating such improvement, regardless of the selection method.

\begin{figure}[hbt]
    \begin{minipage}[t]{0.38\linewidth}
        \centering
        \includegraphics[width=\linewidth]{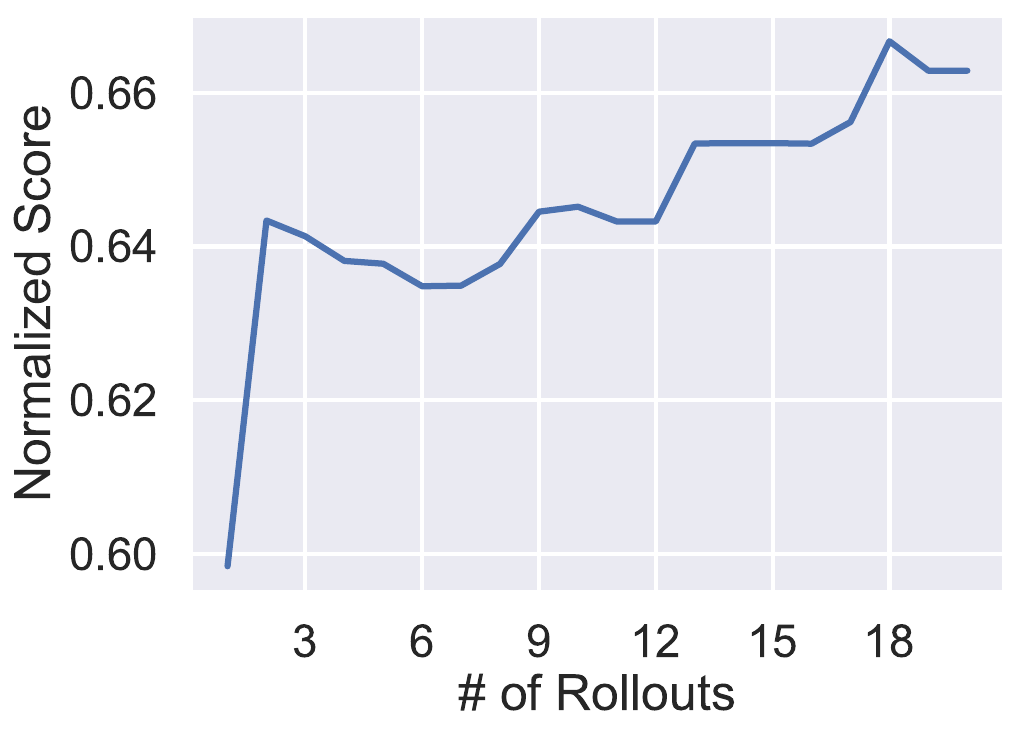}
        \caption{The average performance of \methodname{} on six selected datasets with an increasing number of rollouts.}
        \label{fig:curve}
    \end{minipage}
    \hfill
    \begin{minipage}[t]{0.58\linewidth}
        \centering
        \includegraphics[width=\linewidth]{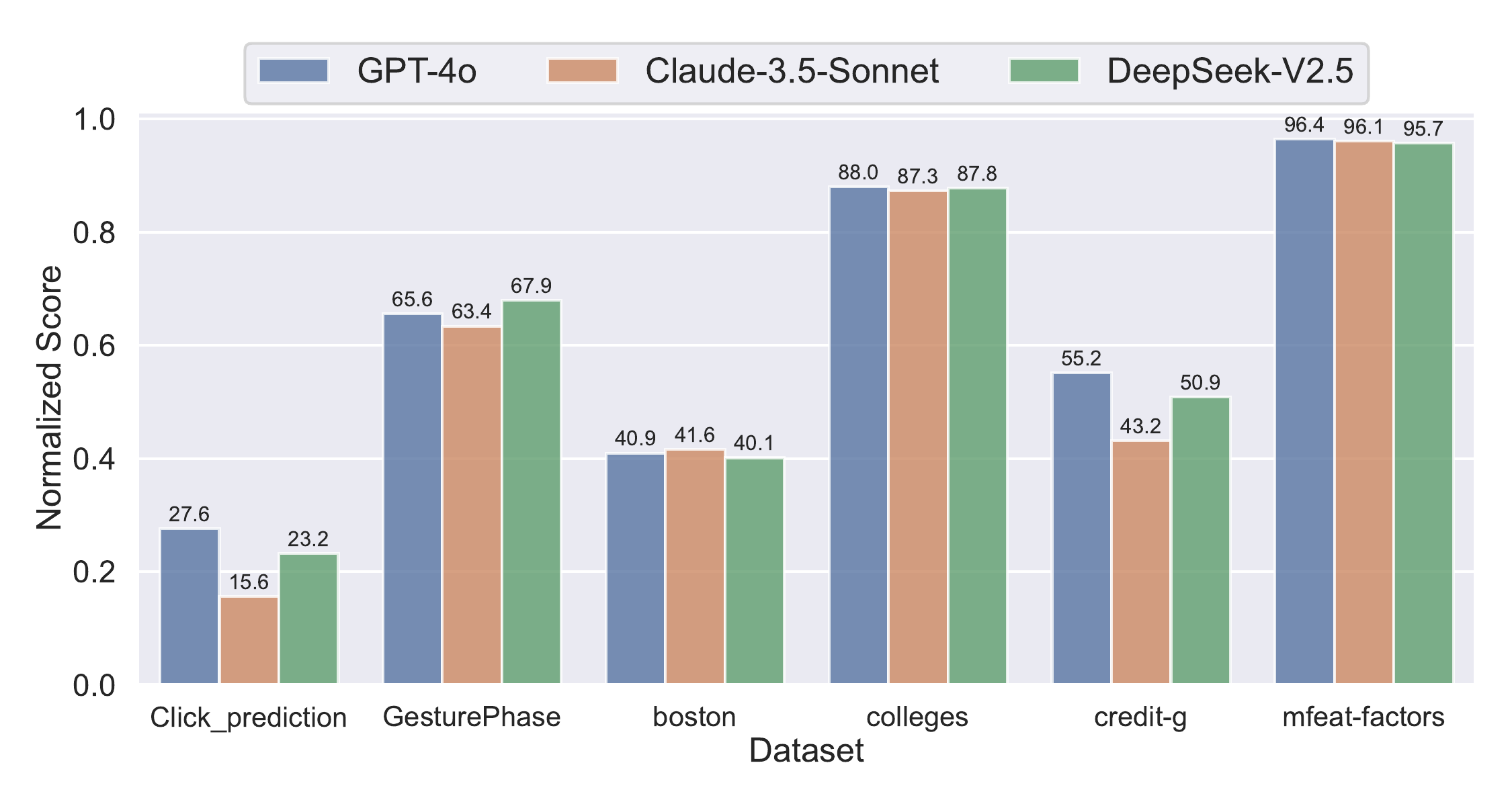}
        \caption{Comparison of Normalized Scores between different base LLMs on six selected datasets.}
        \label{fig:llm}
    \end{minipage}
\end{figure}

\paragraph{Number of Rollouts}
Figure \ref{fig:curve} illustrates that the average performance of \methodname{} improves as the number of permitted rollouts increases. The trend demonstrates the strong scalability of \methodname{}, as it efficiently leverages additional opportunities to explore the search space, improving the normalized score by 4.7\% after 10 rollouts and 6.4\% after 20, compared to the initial rollout.




\paragraph{LLM Adaptability}
To evaluate the robustness of our framework, we conduct experiments using different Large Language Models (LLMs). Specifically, we compare the performance of \methodname{} with \texttt{Claude-3.5-Sonnet} \citep{anthropicIntroducingClaude} and \texttt{GPT-4o} \citep{openaigpt4o} against \texttt{DeepSeek V2.5} which we primarily use for evaluation. This comparison enables us to assess how the choice of LLM affects the overall effectiveness of our approach. 

As Figure \ref{fig:llm} shown, \methodname{} delivers similar results across different LLMs, indicating its flexibility with various models depending on user preference and availability. We also report the numeric results in Appendix \ref{appendix:llms}.


        


\section{Conclusion}

In this paper, we introduced \methodname{}, a novel framework that integrates LLM-based agents with Monte Carlo Tree Search (MCTS) to automate machine learning workflows. Our experimental results, conducted on 20 machine learning datasets, demonstrate \methodname{}’s effectiveness and highlight its distinct advantages over both traditional AutoML frameworks and existing LLM-based approaches. The proposed methodology is not limited to machine learning but could be adapted to a wide range of sequential decision-making problems, provided they can be represented as tree structures with scalar rewards derived from their leaf nodes.
Looking ahead, future work could explore extending this framework to other domains, including software engineering, scientific discovery, game playing, and robotics. Furthermore, improving the efficiency and scalability of the tree search process for larger solution spaces remains an important area for investigation. Another promising direction is developing techniques to provide interpretable explanations of the search process and solution rationale, enhancing the transparency and trustworthiness of the system. \methodname{} represents a significant advancement in automated machine learning, demonstrating the potential of combining traditional search algorithms with the flexibility of LLMs.







\bibliography{references}

\begin{thebibliography}{39}
\providecommand{\natexlab}[1]{#1}
\providecommand{\url}[1]{\texttt{#1}}
\expandafter\ifx\csname urlstyle\endcsname\relax
  \providecommand{\doi}[1]{doi: #1}\else
  \providecommand{\doi}{doi: \begingroup \urlstyle{rm}\Url}\fi

\bibitem[Anthropic(2024)]{anthropicIntroducingClaude}
Anthropic.
\newblock {I}ntroducing {C}laude 3.5 {S}onnet --- anthropic.com.
\newblock \url{https://www.anthropic.com/news/claude-3-5-sonnet}, 2024.

\bibitem[Best et~al.(2019)Best, Cliff, Patten, Mettu, and Fitch]{graeme2019}
Graeme Best, Oliver~M Cliff, Timothy Patten, Ramgopal~R Mettu, and Robert Fitch.
\newblock Dec-mcts: Decentralized planning for multi-robot active perception.
\newblock \emph{The International Journal of Robotics Research}, 38\penalty0 (2-3):\penalty0 316--337, 2019.
\newblock \doi{10.1177/0278364918755924}.

\bibitem[Chi et~al.(2024)Chi, Yang, and Klein]{chi2024thoughtsculptreasoningintermediaterevision}
Yizhou Chi, Kevin Yang, and Dan Klein.
\newblock Thoughtsculpt: Reasoning with intermediate revision and search, 2024.

\bibitem[Clary et~al.(2018)Clary, Morais, Fern, and Hurst]{clary2018}
Patrick Clary, Pedro Morais, Alan Fern, and Jonathan Hurst.
\newblock Monte-carlo planning for agile legged locomotion.
\newblock \emph{Proceedings of the International Conference on Automated Planning and Scheduling}, 28\penalty0 (1):\penalty0 446--450, Jun. 2018.
\newblock \doi{10.1609/icaps.v28i1.13933}.

\bibitem[Coulom(2007)]{MCTS}
R{\'e}mi Coulom.
\newblock Efficient selectivity and backup operators in monte-carlo tree search.
\newblock In H.~Jaap van~den Herik, Paolo Ciancarini, and H.~H. L. M.~(Jeroen) Donkers (eds.), \emph{Computers and Games}, pp.\  72--83, Berlin, Heidelberg, 2007. Springer Berlin Heidelberg.
\newblock ISBN 978-3-540-75538-8.

\bibitem[DeepSeek-AI(2024)]{deepseekv2}
DeepSeek-AI.
\newblock Deepseek-v2: A strong, economical, and efficient mixture-of-experts language model, 2024.

\bibitem[Dwaracherla et~al.(2024)Dwaracherla, Asghari, Hao, and Roy]{dwaracherla2024efficientexplorationllms}
Vikranth Dwaracherla, Seyed~Mohammad Asghari, Botao Hao, and Benjamin~Van Roy.
\newblock Efficient exploration for llms, 2024.

\bibitem[Erickson et~al.(2020)Erickson, Mueller, Shirkov, Zhang, Larroy, Li, and Smola]{erickson2020autogluontabularrobustaccurateautoml}
Nick Erickson, Jonas Mueller, Alexander Shirkov, Hang Zhang, Pedro Larroy, Mu~Li, and Alexander Smola.
\newblock Autogluon-tabular: Robust and accurate automl for structured data, 2020.

\bibitem[Feng et~al.(2023)Feng, Wan, Wen, Wen, Zhang, and Wang]{feng2023alphazero}
Xidong Feng, Ziyu Wan, Muning Wen, Ying Wen, Weinan Zhang, and Jun Wang.
\newblock Alphazero-like tree-search can guide large language model decoding and training, 2023.

\bibitem[Feurer et~al.(2015)Feurer, Klein, Eggensperger, Springenberg, Blum, and Hutter]{feurer-neurips15a}
Matthias Feurer, Aaron Klein, Katharina Eggensperger, Jost Springenberg, Manuel Blum, and Frank Hutter.
\newblock Efficient and robust automated machine learning.
\newblock In \emph{Advances in Neural Information Processing Systems 28 (2015)}, pp.\  2962--2970, 2015.

\bibitem[Feurer et~al.(2020)Feurer, Eggensperger, Falkner, Lindauer, and Hutter]{feurer-arxiv20a}
Matthias Feurer, Katharina Eggensperger, Stefan Falkner, Marius Lindauer, and Frank Hutter.
\newblock Auto-sklearn 2.0: Hands-free automl via meta-learning, 2020.

\bibitem[Gijsbers et~al.(2024)Gijsbers, Bueno, Coors, LeDell, Poirier, Thomas, Bischl, and Vanschoren]{JMLR:v25:22-0493}
Pieter Gijsbers, Marcos L.~P. Bueno, Stefan Coors, Erin LeDell, S{{\'e}}bastien Poirier, Janek Thomas, Bernd Bischl, and Joaquin Vanschoren.
\newblock Amlb: an automl benchmark.
\newblock \emph{Journal of Machine Learning Research}, 25\penalty0 (101):\penalty0 1--65, 2024.

\bibitem[Guo et~al.(2024)Guo, Deng, Wen, Chen, Chang, and Wang]{guo2024dsagentautomateddatascience}
Siyuan Guo, Cheng Deng, Ying Wen, Hechang Chen, Yi~Chang, and Jun Wang.
\newblock Ds-agent: Automated data science by empowering large language models with case-based reasoning, 2024.

\bibitem[Hollmann et~al.(2024)Hollmann, M{\"u}ller, and Hutter]{hollmann2024large}
Noah Hollmann, Samuel M{\"u}ller, and Frank Hutter.
\newblock Large language models for automated data science: Introducing caafe for context-aware automated feature engineering, 2024.

\bibitem[Hong et~al.(2024{\natexlab{a}})Hong, Lin, Liu, Liu, Wu, Li, Chen, Zhang, Wang, Zhang, Zhang, Yang, Zhuge, Guo, Zhou, Tao, Wang, Tang, Lu, Zheng, Liang, Fei, Cheng, Xu, and Wu]{hong2024datainterpreterllmagent}
Sirui Hong, Yizhang Lin, Bang Liu, Bangbang Liu, Binhao Wu, Danyang Li, Jiaqi Chen, Jiayi Zhang, Jinlin Wang, Li~Zhang, Lingyao Zhang, Min Yang, Mingchen Zhuge, Taicheng Guo, Tuo Zhou, Wei Tao, Wenyi Wang, Xiangru Tang, Xiangtao Lu, Xiawu Zheng, Xinbing Liang, Yaying Fei, Yuheng Cheng, Zongze Xu, and Chenglin Wu.
\newblock Data interpreter: An llm agent for data science, 2024{\natexlab{a}}.

\bibitem[Hong et~al.(2024{\natexlab{b}})Hong, Zhuge, Chen, Zheng, Cheng, Wang, Zhang, Wang, Yau, Lin, Zhou, Ran, Xiao, Wu, and Schmidhuber]{hong2024metagpt}
Sirui Hong, Mingchen Zhuge, Jonathan Chen, Xiawu Zheng, Yuheng Cheng, Jinlin Wang, Ceyao Zhang, Zili Wang, Steven Ka~Shing Yau, Zijuan Lin, Liyang Zhou, Chenyu Ran, Lingfeng Xiao, Chenglin Wu, and J{\"u}rgen Schmidhuber.
\newblock Meta{GPT}: Meta programming for a multi-agent collaborative framework.
\newblock In \emph{The Twelfth International Conference on Learning Representations}, 2024{\natexlab{b}}.

\bibitem[Hui \& Tu(2024)Hui and Tu]{hui2024rotenhancinglargelanguage}
Wenyang Hui and Kewei Tu.
\newblock Rot: Enhancing large language models with reflection on search trees, 2024.

\bibitem[Jin et~al.(2019)Jin, Song, and Hu]{jin2019auto}
Haifeng Jin, Qingquan Song, and Xia Hu.
\newblock Auto-keras: An efficient neural architecture search system.
\newblock In \emph{Proceedings of the 25th ACM SIGKDD international conference on knowledge discovery \& data mining}, pp.\  1946--1956, 2019.

\bibitem[Jin et~al.(2023)Jin, Chollet, Song, and Hu]{jin2023autokeras}
Haifeng Jin, Fran{\c{c}}ois Chollet, Qingquan Song, and Xia Hu.
\newblock Autokeras: An automl library for deep learning.
\newblock \emph{Journal of machine Learning research}, 24\penalty0 (6):\penalty0 1--6, 2023.

\bibitem[Kocsis \& Szepesv{\'a}ri(2006)Kocsis and Szepesv{\'a}ri]{10.1007/11871842_29}
Levente Kocsis and Csaba Szepesv{\'a}ri.
\newblock Bandit based monte-carlo planning.
\newblock In Johannes F{\"u}rnkranz, Tobias Scheffer, and Myra Spiliopoulou (eds.), \emph{Machine Learning: ECML 2006}, pp.\  282--293, Berlin, Heidelberg, 2006. Springer Berlin Heidelberg.
\newblock ISBN 978-3-540-46056-5.

\bibitem[Krishnamurthy et~al.(2024)Krishnamurthy, Harris, Foster, Zhang, and Slivkins]{krishnamurthy2024largelanguagemodelsexplore}
Akshay Krishnamurthy, Keegan Harris, Dylan~J. Foster, Cyril Zhang, and Aleksandrs Slivkins.
\newblock Can large language models explore in-context?, 2024.

\bibitem[LeDell \& Poirier(2020)LeDell and Poirier]{H2OAutoML20}
Erin LeDell and Sebastien Poirier.
\newblock {H2O} {A}uto{ML}: Scalable automatic machine learning.
\newblock \emph{7th ICML Workshop on Automated Machine Learning (AutoML)}, July 2020.

\bibitem[Li et~al.(2024)Li, Tan, and Liu]{li2024exploringlargelanguagemodels}
Dawei Li, Zhen Tan, and Huan Liu.
\newblock Exploring large language models for feature selection: A data-centric perspective, 2024.

\bibitem[Liu et~al.(2024)Liu, Gao, and Li]{liu2024large}
Siyi Liu, Chen Gao, and Yong Li.
\newblock Large language model agent for hyper-parameter optimization.
\newblock \emph{arXiv preprint arXiv:2402.01881}, 2024.

\bibitem[Luo et~al.(2024)Luo, Feng, Nong, and Shen]{luo2024autom3l}
Daqin Luo, Chengjian Feng, Yuxuan Nong, and Yiqing Shen.
\newblock Autom3l: An automated multimodal machine learning framework with large language models.
\newblock \emph{arXiv preprint arXiv:2408.00665}, 2024.

\bibitem[Olson \& Moore(2016)Olson and Moore]{olson2016tpot}
Randal~S Olson and Jason~H Moore.
\newblock Tpot: A tree-based pipeline optimization tool for automating machine learning.
\newblock In \emph{Workshop on automatic machine learning}, pp.\  66--74. PMLR, 2016.

\bibitem[OpenAI(2024)]{openaigpt4o}
OpenAI.
\newblock {H}ello {G}{P}{T}-4o.
\newblock \url{https://openai.com/index/hello-gpt-4o/}, 2024.

\bibitem[Schmidt et~al.(2024)Schmidt, Wu, and Jiang]{Schmidt_Wu_Jiang}
Dominik Schmidt, Yuxiang Wu, and Zhengyao Jiang.
\newblock Aide: Human-level performance in data science competitions, 2024.
\newblock URL \url{https://www.weco.ai/blog/technical-report}.

\bibitem[Segler et~al.(2018)Segler, Preuss, and Waller]{segler2018}
Marwin Segler, Mike Preuss, and Mark Waller.
\newblock Planning chemical syntheses with deep neural networks and symbolic ai.
\newblock \emph{Nature}, 555:\penalty0 604--610, 03 2018.
\newblock \doi{10.1038/nature25978}.

\bibitem[Silver et~al.(2016)Silver, Huang, Maddison, Guez, Sifre, van~den Driessche, Schrittwieser, Antonoglou, Panneershelvam, Lanctot, Dieleman, Grewe, Nham, Kalchbrenner, Sutskever, Lillicrap, Leach, Kavukcuoglu, Graepel, and Hassabis]{Silver2016MasteringTG}
David Silver, Aja Huang, Chris~J. Maddison, Arthur Guez, L.~Sifre, George van~den Driessche, Julian Schrittwieser, Ioannis Antonoglou, Vedavyas Panneershelvam, Marc Lanctot, Sander Dieleman, Dominik Grewe, John Nham, Nal Kalchbrenner, Ilya Sutskever, Timothy~P. Lillicrap, Madeleine Leach, Koray Kavukcuoglu, Thore Graepel, and Demis Hassabis.
\newblock Mastering the game of go with deep neural networks and tree search.
\newblock \emph{Nature}, 2016.

\bibitem[Silver et~al.(2017)Silver, Schrittwieser, Simonyan, Antonoglou, Huang, Guez, Hubert, baker, Lai, Bolton, Chen, Lillicrap, Hui, Sifre, van~den Driessche, Graepel, and Hassabis]{Silver2017MasteringTG}
David Silver, Julian Schrittwieser, Karen Simonyan, Ioannis Antonoglou, Aja Huang, Arthur Guez, Thomas Hubert, Lucas baker, Matthew Lai, Adrian Bolton, Yutian Chen, Timothy~P. Lillicrap, Fan Hui, L.~Sifre, George van~den Driessche, Thore Graepel, and Demis Hassabis.
\newblock Mastering the game of go without human knowledge.
\newblock \emph{Nature}, 2017.

\bibitem[Tang et~al.(2024{\natexlab{a}})Tang, Hu, Zhou, Zhong, Zheng, Si, and Ellis]{tang2024coderepairllmsgives}
Hao Tang, Keya Hu, Jin~Peng Zhou, Sicheng Zhong, Wei-Long Zheng, Xujie Si, and Kevin Ellis.
\newblock Code repair with llms gives an exploration-exploitation tradeoff, 2024{\natexlab{a}}.

\bibitem[Tang et~al.(2024{\natexlab{b}})Tang, Fang, Zhou, Yang, Zhong, Hu, Kirchhoff, and Karypis]{tang2024autogluon}
Zhiqiang Tang, Haoyang Fang, Su~Zhou, Taojiannan Yang, Zihan Zhong, Tony Hu, Katrin Kirchhoff, and George Karypis.
\newblock Autogluon-multimodal (automm): Supercharging multimodal automl with foundation models.
\newblock \emph{arXiv preprint arXiv:2404.16233}, 2024{\natexlab{b}}.

\bibitem[Thornton et~al.(2013)Thornton, Hutter, Hoos, and Leyton-Brown]{thornton2013auto}
Chris Thornton, Frank Hutter, Holger~H Hoos, and Kevin Leyton-Brown.
\newblock Auto-weka: Combined selection and hyperparameter optimization of classification algorithms.
\newblock In \emph{Proceedings of the 19th ACM SIGKDD international conference on Knowledge discovery and data mining}, pp.\  847--855, 2013.

\bibitem[Wang et~al.(2024)Wang, Song, Tian, Peng, Yu, Mi, Su, and Yu]{wang2024litesearchefficacioustreesearch}
Ante Wang, Linfeng Song, Ye~Tian, Baolin Peng, Dian Yu, Haitao Mi, Jinsong Su, and Dong Yu.
\newblock Litesearch: Efficacious tree search for llm, 2024.

\bibitem[Wang et~al.(2021)Wang, Wu, Weimer, and Zhu]{wang2021flaml}
Chi Wang, Qingyun Wu, Markus Weimer, and Erkang Zhu.
\newblock Flaml: A fast and lightweight automl library.
\newblock In \emph{MLSys}, 2021.

\bibitem[Wu et~al.(2015)Wu, Ramchurn, Jiang, Fischer, Rodden, and Jennings]{wu2015}
Feng Wu, Sarvapali~D. Ramchurn, Wenchao Jiang, Jeol~E. Fischer, Tom Rodden, and Nicholas~R. Jennings.
\newblock Agile planning for real-world disaster response.
\newblock In \emph{Proceedings of the 24th International Conference on Artificial Intelligence}, IJCAI'15, pp.\  132–138. AAAI Press, 2015.
\newblock ISBN 9781577357384.

\bibitem[Zhang et~al.(2024)Zhang, Zhang, Ren, Li, and Yang]{zhang2024mlcopilotunleashingpowerlarge}
Lei Zhang, Yuge Zhang, Kan Ren, Dongsheng Li, and Yuqing Yang.
\newblock Mlcopilot: Unleashing the power of large language models in solving machine learning tasks, 2024.

\bibitem[Zhou et~al.(2024)Zhou, Yan, Shlapentokh-Rothman, Wang, and Wang]{zhou2024languageagenttreesearch}
Andy Zhou, Kai Yan, Michal Shlapentokh-Rothman, Haohan Wang, and Yu-Xiong Wang.
\newblock Language agent tree search unifies reasoning acting and planning in language models, 2024.

\end{thebibliography}
\bibliographystyle{iclr2025_conference}

\appendix
\newpage
\newpage
\section{Datasets}

Table \ref{table:datasets} outlines the detailed information of the datasets used for evaluation.

\begin{table}[H]
\centering
\small
\resizebox{.99\textwidth}{!}{
\begin{tabular}{>{\scriptsize}p{0.25\textwidth}>{\scriptsize}p{0.1\textwidth}<{\centering}>{\scriptsize}c>{\scriptsize}c>{\scriptsize}c>{\scriptsize}c>{\scriptsize}c}

\toprule
\footnotesize \textbf{Dataset name}    & \footnotesize \textbf{\# Features} & \footnotesize \textbf{\# Rows}  & \footnotesize \textbf{\# Classes} & \footnotesize \textbf{Task Type}  &  \footnotesize \textbf{Metric}  & \footnotesize \textbf{Source}   \\
\midrule
boston          & 14          & 506      & N/A        & Regression & RMSE  & OpenML \scriptsize (Dataset ID: 531)     \\
colleges                                  & 48          & 7063     & N/A        & Regression & RMSE   & OpenML \scriptsize (Dataset ID: 42727)     \\
concrete-strength        & 9           & 4866     & N/A        & Regression & RMSE  & Kaggle \scriptsize (playground-series-s3e9)      \\
diamonds                                  & 10          & 53940    & N/A        & Regression & RMSE  & OpenML \scriptsize (Dataset ID: 42225)     \\
house-prices                               & 81          & 1460     & N/A        & Regression & RMSE & Kaggle  \scriptsize (house-prices-advanced-regression-techniques)     \\
Moneyball                                 & 15          & 1232     & N/A        & Regression & RMSE  & OpenML \scriptsize (Dataset ID: 41021)      \\
SAT11-HAND-runtime-regression             & 118         & 4440     & N/A        & Regression & RMSE  & OpenML \scriptsize (Dataset ID: 41980)    \\
credit-g                                  & 21          & 1000     & 2          & Classification & F1 & OpenML \scriptsize (Dataset ID: 31)      \\
Click\_prediction\_small                    & 12          & 39948    & 2          & Classification & F1  & OpenML \scriptsize (Dataset ID: 42733)    \\

icr   & 58          & 617      & 2          & Classification & F1 & Kaggle  \scriptsize (icr-identify-age-related-conditions)     \\
jasmine                                   & 145         & 2984     & 2          & Classification & F1 & OpenML \scriptsize (Dataset ID: 41143)       \\
kc1                                       & 21          & 2109     & 2          & Classification & F1  & OpenML \scriptsize (Dataset ID: 1067)     \\
kick                                      & 33          & 72983    & 2          & Classification & F1  & OpenML \scriptsize (Dataset ID: 41162)     \\
smoker-status                             & 23          & 143330   & 2          & Classification & F1  & Kaggle \scriptsize (playground-series-s3e24)     \\
software-defects                          & 22          & 91586    & 2          & Classification & F1  & Kaggle \scriptsize (playground-series-s3e23)     \\
titanic                                   & 12          & 891      & 2          & Classification & F1  & Kaggle \scriptsize (titanic)    \\
GesturePhaseSegmentationProcessed      & 33          & 9873     & 5          & Multiclass & F1-weighted & OpenML \scriptsize (Dataset ID: 4538) \\

mfeat-factors                             & 217         & 2000     & 10         & Multiclass & F1-weighted  & OpenML \scriptsize (Dataset ID: 12)\\

segment                                   & 20          & 2310     & 7          & Multiclass & F1-weighted & OpenML \scriptsize (Dataset ID: 40984) \\

wine-quality-white                      & 12          & 4898     & 7          & Multiclass & F1-weighted & OpenML \scriptsize (Dataset ID: 40498) \\
\bottomrule
\hline
\end{tabular}
}
\caption{Summary of the machine learning datasets used in the experiments. OpenML datasets can be accessed using their respective dataset IDs. The Kaggle datasets are available at https://www.kaggle.com/competitions/\{source\}.}
\label{table:datasets}
\end{table}

\newpage
\section{Prompts}

\subsection{Task Prompt}
All LLM-based methods start by receiving the same base requirement prompt at the beginning of the task. The prompt specifies the dataset's name, the target label column, the evaluation metric to be used, and the dataset's file path. Furthermore, the prompt include a path to a text file containing the dataset's metadata.

\begin{lstlisting}[style=pythonstyle]
TASK_PROMPT = """
# User requirement
This is a {datasetname} dataset. 
Your goal is to predict the target column `{target_col}`.
Perform data analysis, data preprocessing, feature engineering, and modeling to predict the target. Report {metric} on the eval data. Do not plot or make any visualizations.

# Data dir
train set (with labels): {train_path}
dev set (with labels): {dev_path}
test set (without labels): {test_path}
dataset description: {data_info_path} 
(During EDA, you can use this file 
to get additional information about the dataset)
"""
\end{lstlisting}

Since AIDE automatically splits the training data into a new train set and a validation set, we combine the original train and validation sets and provide them as input to AIDE. We set $\text{k\_fold\_validation}$ to 1 in its configuration file to enforce a single train-val split for closer alignment with our setup. In both setups, the frameworks have access to the labels for both the train and validation sets.

\subsection{Instruction Prompt}
The instruction prompt would direct the framework to save the final prediction file for evaluation.

\begin{lstlisting}[style=pythonstyle]
DI_INSTRUCTION = """
## Attention
1. Please do not leak the target label in any form during training.
2. Test set does not have the target column.
3. When conducting data exploration or analysis, print out the results of your findings.
4. You should perform transformations on train, dev, and test sets at the same time (it's a good idea to define functions for this and avoid code repetition).
5. When scaling or transforming features, make sure the target column is not included.
6. You could utilize dev set to validate and improve model training. {special_instruction}

## Saving Dev and Test Predictions
1. Save the prediction results of BOTH the dev set and test set in `dev_predictions.csv` and `test_predictions.csv` respectively in the output directory. 
- Both files should contain a single column named `target` with the predicted values.
2. Make sure the prediction results are in the same format as the target column in the training set. 
- For instance, if the target column is categorical, the prediction results should be categorical as well.

## Output Performance
Print the train and dev set performance in the last step.

# Output dir
{output_dir}
"""
\end{lstlisting}

\newpage
\subsection{Insight Proposal Prompt}
Insight Proposer uses this prompt to generate a search space of insights for different stages of the machine learning task.
\begin{lstlisting}[style=pythonstyle]
DATASET_INSIGHT_PROMPT = """
# Dataset Description
{dataset}

# Dataset Metadata
{metadata}

# Dataset Head
{head}

# Instruction
Propose insights to help improve the performance of the model on this dataset.
The insights should be proposed based on the dataset description with different task types.
Each task type should have at least 5 insights.
Make sure each method is diverse enough and can be implemented separately.
Be specific about models' choices, ensemble and tuning techniques, and preprocessing & feature engineering techniques.

# Format
```json
[
    {{
        "task_type": "EDA",
        "insights": [
            "insight1",
            "insight2",
            "insight3",
            ...
            "insightN"
        ]   
    }},
    {{
        "task_type": "Data Preprocessing",
        "insights": [
            "insight1",
            "insight2",
            "insight3",
            ...
            "insightN"
        ]   
    }},
    {{
        "task_type": "Feature Engineering",
        "insights": [
            "insight1",
            "insight2",
            "insight3",
            ...
            "insightN"
        ]   
    }},
    {{
        "task_type": "Model Training",
        "insights": [
            "insight1",
            "insight2",
            "insight3",
            ...
            "insightN"
        ]   
    }}
]
```
"""
\end{lstlisting}

\newpage
\section{Results}
\label{appendix:results}
\subsection{Main Results}

\begin{table}[hbt]
\centering
\resizebox{0.99\textwidth}{!}{
\begin{tabularx}{1.1\textwidth}{>{\scriptsize}lcccccccccc}
\toprule
 & \multicolumn{2}{c}{\textbf{AutoGluon}} & \multicolumn{2}{c}{\textbf{AutoSklearn}} & \multicolumn{2}{c}{\textbf{AIDE}} & \multicolumn{2}{c}{\textbf{DI}} & \multicolumn{2}{c}{\textbf{\methodname{}}}  \\ 
\normalsize Dataset & {Avg.} & {Best} & {Avg.} & {Best} & {Avg.} & {Best} & {Avg.} & {Best} & {Avg.} & {Best} \\
\midrule
Click\_prediction\_small & 7 & 7 & 2 & 1 & 7.3 & 4 & 11 & 10 & 7.7 & 6 \\
GesturePhaseSegmentationProcessed & 1 & 1 & 6.3 & 3 & 7.3 & 4 & 11 & 10 & 5.3 & 2 \\
Moneyball & 4 & 4 & 10 & 9 & 4 & 1 & 9 & 2 & 6 & 3 \\
SAT11-HAND-runtime-regression & 1 & 1 & 12 & 11 & 5.3 & 3 & 9 & 8 & 3.7 & 2 \\
boston & 5 & 5 & 12 & 11 & 3.7 & 2 & 9 & 8 & 4 & 1 \\
colleges & 1 & 1 & 12 & 11 & 6 & 2 & 8 & 7 & 4 & 3 \\
concrete-strength & 5 & 5 & 12 & 11 & 6.3 & 4 & 2 & 1 & 8.3 & 6 \\
credit-g & 4 & 4 & 10 & 9 & 10 & 5 & 5.3 & 1 & 3.7 & 2 \\
diamonds & 2 & 2 & 12 & 11 & 6 & 4 & 8.7 & 7 & 3 & 1 \\
house-prices & 1 & 1 & 12 & 11 & 6.7 & 5 & 7.3 & 3 & 4 & 2 \\
icr & 5 & 5 & 5.3 & 3 & 12 & 11 & 9 & 8 & 2.3 & 1 \\
jasmine & 7 & 7 & 6 & 4 & 8.7 & 5 & 11.3 & 9 & 2 & 1 \\
kc1 & 10 & 10 & 2.7 & 1 & 8 & 5 & 11.3 & 9 & 5 & 2 \\
kick & 4 & 4 & 2 & 1 & 9.3 & 6 & 11 & 10 & 6.7 & 5 \\
mfeat-factors & 4 & 4 & 2 & 1 & 10 & 9 & 10.3 & 6 & 6.7 & 5 \\
segment & 3 & 3 & 6.3 & 5 & 11 & 10 & 9.7 & 7 & 2.3 & 1 \\
smoker-status & 7 & 7 & 4.7 & 3 & 11.3 & 9 & 7.7 & 2 & 4.3 & 1 \\
software-defects & 8 & 8 & 2 & 1 & 12 & 11 & 6 & 4 & 7.7 & 6 \\
titanic & 7 & 7 & 9.7 & 6 & 2.7 & 1 & 10.3 & 8 & 5.3 & 3 \\
wine-quality-white & 2 & 2 & 10 & 8 & 7.3 & 4 & 9 & 7 & 3.3 & 1 \\
\midrule 

\normalsize Overall Rank $\downarrow$ & \textbf{4.4} & 4.4  & 7.6 & 6.1 & 7.8 & 5.3 & 8.8 & 6.4 & 4.8 &\textbf{2.7} \\
\bottomrule
\end{tabularx}
}
\caption{Methods' ranking for each tabular dataset}
\label{main-table}
\end{table}

\begin{table}[H]
\centering
\resizebox{0.99\textwidth}{!}{
\begin{tabularx}{1.1\textwidth}{>{\scriptsize}lcccccccccc}
\toprule
 & \multicolumn{2}{c}{\textbf{AutoGluon}} & \multicolumn{2}{c}{\textbf{AutoSklearn}} & \multicolumn{2}{c}{\textbf{AIDE}} & \multicolumn{2}{c}{\textbf{DI}} & \multicolumn{2}{c}{\textbf{\methodname{}}}  \\ 
\normalsize Dataset & {Avg.} & {Best} & {Avg.} & {Best} & {Avg.} & {Best} & {Avg.} & {Best} & {Avg.} & {Best} \\
\midrule
Click\_prediction\_small & 26.6 & 26.6 & 40.2 & 40.3 & 26.1 & 39.4 & 12.9 & 13.9 & 23.2 & 27.4 \\
GesturePhaseSegmentationProcessed & 69.3 & 69.3 & 67.2 & 68.4 & 56.3 & 68.1 & 60.1 & 64.4 & 67.9 & 69.2 \\
Moneyball & 24.3 & 24.3 & 13.1 & 13.8 & 23.8 & 24.6 & 9.5 & 24.5 & 21.9 & 24.5 \\
SAT11-HAND-runtime-regression & 12.6 & 12.6 & 10.3 & 10.3 & 12.0 & 12.1 & 11.4 & 11.9 & 12.2 & 12.5 \\
boston & 39.8 & 39.8 & 19.5 & 19.6 & 40.5 & 41.3 & 37.0 & 38.6 & 40.1 & 41.4 \\
colleges & 88.3 & 88.3 & 2.1 & 2.1 & 86.0 & 87.8 & 87.5 & 87.7 & 87.8 & 87.8 \\
concrete-strength & 28.3 & 28.3 & 17.4 & 17.9 & 28.3 & 28.3 & 28.8 & 29.6 & 28.2 & 28.2 \\
credit-g & 50.5 & 50.5 & 35.1 & 44.0 & 21.6 & 48.4 & 48.1 & 53.2 & 50.9 & 52.7 \\
diamonds & 13.8 & 13.8 & 8.7 & 8.7 & 13.7 & 13.7 & 13.5 & 13.6 & 13.7 & 13.8 \\
house-prices & 9.0 & 9.0 & 2.0 & 2.0 & 8.9 & 8.9 & 8.5 & 9.0 & 8.9 & 9.0 \\
icr & 76.2 & 76.2 & 70.4 & 79.2 & 31.7 & 35.9 & 57.8 & 60.6 & 78.7 & 79.2 \\
jasmine & 84.3 & 84.3 & 84.4 & 84.7 & 83.6 & 84.6 & 77.8 & 83.5 & 85.4 & 86.2 \\
kc1 & 38.3 & 38.3 & 43.5 & 45.0 & 40.8 & 42.6 & 38.1 & 41.2 & 42.2 & 43.1 \\
kick & 39.6 & 39.6 & 41.8 & 42.1 & 14.9 & 38.6 & 2.8 & 4.2 & 35.9 & 38.7 \\
mfeat-factors & 96.7 & 96.7 & 97.1 & 97.5 & 94.4 & 94.5 & 93.0 & 96.0 & 95.7 & 96.2 \\
segment & 93.5 & 93.5 & 92.7 & 93.1 & 91.7 & 92.2 & 91.7 & 92.6 & 93.8 & 94.4 \\
smoker-status & 78.0 & 78.0 & 78.6 & 78.9 & 74.8 & 76.3 & 77.3 & 81.5 & 82.4 & 91.5 \\
software-defects & 51.5 & 51.5 & 61.1 & 61.7 & 49.7 & 49.8 & 54.5 & 57.3 & 52.2 & 53.3 \\
titanic & 78.9 & 78.9 & 76.2 & 78.9 & 81.2 & 83.7 & 76.0 & 78.5 & 78.8 & 79.7 \\
wine-quality-white & 65.4 & 65.4 & 60.7 & 61.4 & 62.9 & 65.1 & 61.2 & 61.6 & 65.3 & 66.0 \\
\midrule
\normalsize Overall NS \% $\uparrow$ & 53.2 & 53.2 & 46.1 & 47.5  & 45.5 & 51.8 & 47.4 & 50.2 & \textbf{53.3} & \textbf{54.7}  \\
\bottomrule
\end{tabularx}
}
\caption{Methods' NS \% for each tabular dataset}
\label{table:full-main-results}
\end{table}


\subsection{Performance using different LLMs}
\label{appendix:llms}

\begin{table}[hbt]
\centering
\begin{tabular}{lccc}
    \toprule
    & \textbf{GPT-4o} & \textbf{Claude 3.5 Sonnet} & \textbf{DeepSeek V2.5}   \\ \midrule
    Avg. NS $\uparrow$ & \textbf{62.3}   & 57.9    & 60.9 \\
    Avg. Best NS $\uparrow$  & \textbf{65.5} & 59.2 & 62.4 \\
    \midrule
    Avg. Rank $\downarrow$   & \textbf{3.7}  & 6.3     & 5.0    \\
    Avg. Best Rank $\downarrow$  & \textbf{1.5}    & 4.8     & 3.2    \\ 
    \bottomrule
\end{tabular}
\caption{Results of \methodname{} with different base LLMs on the selected tabular datasets.}
\label{table:llm}
\end{table}

\newpage

\section{Cost-effectiveness Analysis}
\label{section:cost}

We conduct multiple trials of execution of each method to estimate the average running cost for the LLM-based baselines. As shown in Table \ref{table:cost}, all methods incur relatively low costs to complete a single machine learning task. Among these, AIDE exhibits the lowest execution cost, due to the lack of stage-wise planning, resulting in fewer token generations compared to the other approaches. Additionally, \methodname{}, which employs Data Interpreter as its base experimenter, is less costly than Data Interpreter itself. This efficiency is largely due to \methodname{}'s state-saving and loading mechanism, which reduces the generation of repeated tasks and code.

\begin{table}[hbt]
    \centering
    \begin{tabular}{lccc}
        \toprule
         & Cost per ML task (\$) & & \\
        \midrule 
        Data Interpreter ($k$=10) & 0.07 & \\
        AIDE ($k$=10) & 0.01 & \\
        \methodname{} ($k$=10) & 0.05 & \\
    \bottomrule
    \end{tabular}
    \caption{Estimated costs of agent-based frameworks utilizing DeepSeekV2.5 on a single machine learning dataset over $k$ iterations/rollouts.}
    \label{table:cost}
\end{table}

\newpage
\section{Case Study}
\subsection{Overview of SELA's search process}
\begin{lstlisting}[style=txtfile]
Number of simulations: 10
[Node 0]
Plans: 
1. Perform exploratory data analysis on the train and dev datasets
2. Preprocess the train, dev, and test datasets
3. Perform feature engineering on the train, dev, and test datasets
4. Train multiple models and evaluate their performance
5. Train a weighted ensemble model using the best performing models
6. Evaluate the ensemble model on the dev set and save predictions
7. Generate predictions for the test set and save them
Simulated: True
Score: avg score: 0.6150206840685731, simulated score: {'train_score': 1.0, 'dev_score': 0.6855841857240594, 'test_score': 0.6814818772150697, 'score': 0.6855841857240594}, Visits: 10

	[Node 0-0]
	Plans: 
	3. Perform feature engineering on the train, dev, and test datasets by creating new features that calculate the magnitude of the vectorial velocities and accelerations to capture the overall movement intensity.
	Simulated: True
	Score: avg score: 0.6507249985568175, simulated score: {'train_score': 0.982920964830782, 'dev_score': 0.6420233166755841, 'test_score': 0.647550336228104, 'score': 0.6420233166755841}, Visits: 2

		[Node 0-0-0]
		Plans: 
		4. Train a Random Forest classifier to leverage its ability to handle high-dimensional data and capture non-linear relationships, and evaluate its performance
		Simulated: False
		Score: avg score: 0, simulated score: {}, Visits: 0

		[Node 0-0-1]
		Plans: 
		4. Train multiple models, including a Support Vector Machine (SVM) with a radial basis function (RBF) kernel, and evaluate their performance.
		Simulated: False
		Score: avg score: 0, simulated score: {}, Visits: 0

		[Node 0-0-2]
		Plans: 
		4. Implement a Neural Network with multiple layers to capture the hierarchical patterns in the data and evaluate its performance
		Simulated: True
		Score: avg score: 0.6594266804380511, simulated score: {'train_score': 1.0, 'dev_score': 0.6594266804380511, 'test_score': 0.6702614538699305, 'score': 0.6594266804380511}, Visits: 1

		[Node 0-0-3]
		Plans: 
		4. Train multiple models, apply an ensemble method like Gradient Boosting to combine them, and evaluate their performance
		Simulated: False
		Score: avg score: 0, simulated score: {}, Visits: 0

		[Node 0-0-4]
		Plans: 
		4. Train multiple models, perform hyperparameter tuning using Grid Search or Random Search, and evaluate their performance
		Simulated: False
		Score: avg score: 0, simulated score: {}, Visits: 0

	[Node 0-1]
	Plans: 
	3. Perform feature engineering on the train, dev, and test datasets by generating time-based features, such as the difference between consecutive frames, to capture the rate of change in movements.
	Simulated: True
	Score: avg score: 0.6464940718972336, simulated score: {'train_score': 1.0, 'dev_score': 0.5985614604756948, 'test_score': 0.5857379626419719, 'score': 0.5985614604756948}, Visits: 2

		[Node 0-1-0]
		Plans: 
		4. Train a Random Forest classifier to leverage its ability to handle high-dimensional data and capture non-linear relationships
		Simulated: False
		Score: avg score: 0, simulated score: {}, Visits: 0

		[Node 0-1-1]
		Plans: 
		4. Train multiple models, including a Support Vector Machine (SVM) with a radial basis function (RBF) kernel, and evaluate their performance to model the complex decision boundaries between different gesture phases.
		Simulated: True
		Score: avg score: 0.6944266833187726, simulated score: {'train_score': 1.0, 'dev_score': 0.6944266833187726, 'test_score': 0.6928451194338062, 'score': 0.6944266833187726}, Visits: 1

		[Node 0-1-2]
		Plans: 
		4. Implement a Neural Network with multiple layers to capture the hierarchical patterns in the data and evaluate its performance
		Simulated: False
		Score: avg score: 0, simulated score: {}, Visits: 0

		[Node 0-1-3]
		Plans: 
		4. Train multiple models, apply an ensemble method like Gradient Boosting to combine them, and evaluate their performance
		Simulated: False
		Score: avg score: 0, simulated score: {}, Visits: 0

		[Node 0-1-4]
		Plans: 
		4. Train multiple models and perform hyperparameter tuning using techniques like Grid Search or Random Search to optimize and evaluate their performance.
		Simulated: False
		Score: avg score: 0, simulated score: {}, Visits: 0

	[Node 0-2]
	Plans: 
	3. Perform feature engineering on the train, dev, and test datasets by creating features that represent the spatial relationships between different body parts, such as the distance between the hands and the head.
	Simulated: True
	Score: avg score: 0.6296836159165489, simulated score: {'train_score': 0.7619969104124632, 'dev_score': 0.5997286931710517, 'test_score': 0.604077566134264, 'score': 0.5997286931710517}, Visits: 3

		[Node 0-2-0]
		Plans: 
		4. Train a Random Forest classifier to leverage its ability to handle high-dimensional data and capture non-linear relationships, and evaluate its performance
		Simulated: False
		Score: avg score: 0, simulated score: {}, Visits: 0

		[Node 0-2-1]
		Plans: 
		4. Train multiple models, including a Support Vector Machine (SVM) with a radial basis function (RBF) kernel, and evaluate their performance to model the complex decision boundaries between different gesture phases.
		Simulated: True
		Score: avg score: 0.6446610772892973, simulated score: {'train_score': 0.9952809245924918, 'dev_score': 0.6372459669415207, 'test_score': 0.6423549137767338, 'score': 0.6372459669415207}, Visits: 2

			[Node 0-2-1-0]
			Plans: 
			5. Train a weighted ensemble model using the best performing models from task 4
			Simulated: False
			Score: avg score: 0, simulated score: {}, Visits: 0

			[Node 0-2-1-1]
			Plans: 
			5. Using the models that performed best in task 4, train a weighted ensemble model to improve overall performance.
			Simulated: False
			Score: avg score: 0, simulated score: {}, Visits: 0

			[Node 0-2-1-2]
			Plans: 
			5. Develop a weighted ensemble model by integrating the top-performing models from task 4, ensuring to evaluate and adjust the weights for optimal performance.
			Simulated: True
			Score: avg score: 0.6520761876370741, simulated score: {'train_score': 1.0, 'dev_score': 0.6520761876370741, 'test_score': 0.6563435152603494, 'score': 0.6520761876370741}, Visits: 1

			[Node 0-2-1-3]
			Plans: 
			5. Train a weighted ensemble model by combining the predictions of the top-performing models from task 4 to improve overall performance.
			Simulated: False
			Score: avg score: 0, simulated score: {}, Visits: 0

			[Node 0-2-1-4]
			Plans: 
			5. Develop a weighted ensemble model by combining the top-performing models from task 4, ensuring to optimize the weights for improved performance.
			Simulated: False
			Score: avg score: 0, simulated score: {}, Visits: 0

		[Node 0-2-2]
		Plans: 
		4. Implement a Neural Network with multiple layers to capture the hierarchical patterns in the data and evaluate its performance
		Simulated: False
		Score: avg score: 0, simulated score: {}, Visits: 0

		[Node 0-2-3]
		Plans: 
		4. Train multiple models, apply an ensemble method like Gradient Boosting to combine them, and evaluate their performance
		Simulated: False
		Score: avg score: 0, simulated score: {}, Visits: 0

		[Node 0-2-4]
		Plans: 
		4. Perform hyperparameter tuning using Grid Search or Random Search to train multiple models and evaluate their performance
		Simulated: False
		Score: avg score: 0, simulated score: {}, Visits: 0

	[Node 0-3]
	Plans: 
	3. Apply feature selection techniques such as Recursive Feature Elimination (RFE) or SelectKBest to identify and retain the most important features in the train, dev, and test datasets.
	Simulated: True
	Score: avg score: 0.49056683315196203, simulated score: {'train_score': 0.9988177730410426, 'dev_score': 0.51620611302976, 'test_score': 0.525989891002361, 'score': 0.51620611302976}, Visits: 2

		[Node 0-3-0]
		Plans: 
		4. Train a Random Forest classifier to leverage its ability to handle high-dimensional data and capture non-linear relationships, and evaluate its performance.
		Simulated: False
		Score: avg score: 0, simulated score: {}, Visits: 0

		[Node 0-3-1]
		Plans: 
		4. Train multiple models, including a Support Vector Machine (SVM) with a radial basis function (RBF) kernel, and evaluate their performance to model the complex decision boundaries between different gesture phases.
		Simulated: True
		Score: avg score: 0.4649275532741641, simulated score: {'train_score': 0.7299159411193588, 'dev_score': 0.4649275532741641, 'test_score': 0.4631598897487413, 'score': 0.4649275532741641}, Visits: 1

		[Node 0-3-2]
		Plans: 
		4. Implement and train a Neural Network with multiple layers to capture hierarchical patterns in the data and evaluate its performance
		Simulated: False
		Score: avg score: 0, simulated score: {}, Visits: 0

		[Node 0-3-3]
		Plans: 
		4. Train multiple models, apply an ensemble method like Gradient Boosting to combine them, and evaluate their performance
		Simulated: False
		Score: avg score: 0, simulated score: {}, Visits: 0

		[Node 0-3-4]
		Plans: 
		4. Train multiple models, perform hyperparameter tuning using techniques like Grid Search or Random Search, and evaluate their performance
		Simulated: False
		Score: avg score: 0, simulated score: {}, Visits: 0

	[Node 0-4]
	Plans: 
	3. Create interaction features by combining existing features, such as the product of velocity and acceleration, to capture complex relationships in the train, dev, and test datasets
	Simulated: False
	Score: avg score: 0, simulated score: {}, Visits: 0

Generated 29 unique codes.
Best node: 0-1-1, score: {'train_score': 1.0, 'dev_score': 0.6944266833187726, 'test_score': 0.6928451194338062, 'score': 0.6944266833187726}
Dev best node: 0-1-1, score: {'train_score': 1.0, 'dev_score': 0.6944266833187726, 'test_score': 0.6928451194338062, 'score': 0.6944266833187726}

\end{lstlisting}

In this case study, we demonstrate how SELA conducts a search cycle using MCTS: 

\textbf{Pre-search Step: Initialization}\\
SELA begins by defining high-level stages, such as exploratory data analysis, data preprocessing, feature engineering, and model training, which structure the overall machine learning workflow. During the search, SELA populates these stages with specific insights, which act as experimental configurations for simulation.

\textbf{Step 1 \& 2: Selection and Expansion}\\
SELA leverages MCTS to explore specific stages like feature engineering and model training. For example, in one iteration, SELA selects Node 0-1. This node corresponds to a stage insight that generates time-based features, expanding into five child nodes representing various model specifications and training strategies, such as Random Forests, Support Vector Machines, Neural Networks, Gradient Boosting, or Grid Search.

\textbf{Step 3: Simulation}\\
Next, SELA samples one of the expanded child nodes for simulation. For instance, when Node 0-1-1 is chosen, SELA runs a complete experiment where time-based feature engineering (Node 0-1) is followed by training a Support Vector Machine (SVM) with a kernel specified by Node 0-1-1. The simulation yields an evaluation score.

\textbf{Step 4: Backpropagation}\\
After the simulation, the resulting performance score is propagated back through the tree. For example, after simulating Node 0-1-1, MCTS updates the numeric feedback for its parent nodes, such as Node 0-1 and Node 0. The search cycle repeats from Steps 1 to 4 until a stopping condition is reached.

\textbf{Post-search Step: Best Node Selection}\\
In the final phase, SELA selects the node representing the best-performing solution. In this example, Node 0-1-1, using an SVM with an RBF kernel, achieved the highest score in the current dataset by combining effective feature engineering with advanced model training. SELA then presents the code associated with this node as the optimal solution.

\end{document}